\documentclass[12pt,letterpaper]{article} 
\usepackage[margin=1.2in]{geometry} 
\usepackage{setspace} 
\usepackage[utf8]{inputenc}
\usepackage{amsmath}
\usepackage{amssymb}
\usepackage{amsfonts}
\usepackage{graphicx} 
\usepackage{hyperref} 
\usepackage{enumitem} 
\usepackage{float} 
\usepackage{booktabs} 
\usepackage{float}
\usepackage{graphicx}

\title{Uncertainty Quantification for Machine Learning-Based Prediction: A Polynomial Chaos Expansion Approach for Joint Model and Input Uncertainty Propagation}
\author{Xiaoping Du\\ \small School of Mechanical Engineering \\ \small Purdue University }
\date{\today} 

\begin{document}

\singlespacing
\maketitle
\doublespacing 

\begin{abstract}
Machine learning (ML) surrogate models are increasingly used in engineering analysis and design to replace computationally expensive simulation models, significantly reducing computational cost and accelerating decision-making processes. However, ML predictions contain inherent errors, often estimated as model uncertainty, which is coupled with variability in model inputs. Accurately quantifying and propagating these combined uncertainties is essential for generating reliable engineering predictions. This paper presents a robust framework based on Polynomial Chaos Expansion (PCE) to handle joint input and model uncertainty propagation. While the approach applies broadly to general ML surrogates, we focus on Gaussian Process regression models, which provide explicit predictive distributions for model uncertainty. By transforming all random inputs into a unified standard space, a PCE surrogate model is constructed, allowing efficient and accurate calculation of the mean and standard deviation of the output. The proposed methodology also offers a mechanism for global sensitivity analysis, enabling the accurate quantification of the individual contributions of input variables and ML model uncertainty to the overall output variability. This approach provides a computationally efficient and interpretable framework for comprehensive uncertainty quantification, supporting trustworthy ML predictions in downstream engineering applications.
\end{abstract}

\section{Introduction and Literature Review}

As engineering systems become increasingly complex, the reliance on high-fidelity computational models for design, analysis, and decision-making has intensified. These simulation models provide detailed insights into system behavior, yet their high computational demands make them impractical for tasks requiring numerous evaluations, such as optimization, uncertainty quantification (UQ), or reliability assessment. To mitigate this computational burden, surrogate modeling has emerged as a key enabler of simulation-based design. Machine learning (ML) surrogates, in particular, have shown significant promise in approximating complex mappings between  inputs and outputs. If the original computational model is denoted by \( Y = h(\mathbf{X}) \), where \( \mathbf{X} \) is a vector of input variables, the surrogate model approximates it as \( Y = g(\mathbf{X}) \), dramatically reducing the evaluation cost \cite{sacks1989design, forrester2008engineering}.

Yet, the deployment of ML surrogates introduces a critical challenge:  \textit{model uncertainty}. ML predictions are inherently imperfect due to limited training data, model misspecification, or extrapolation beyond the sampled design space. Simultaneously, the input variables \( \mathbf{X} \) are rarely known deterministically in real-world applications, introducing \textit{input uncertainty}. These two sources—epistemic (model) and aleatory (input) uncertainty—must be jointly considered to enable informed and robust engineering decisions. Foundational work in structural reliability, such as that presented in \cite{ditlevsen1982model} and \cite{ditlevsen1996structural}, emphasizes the need to handle these coupled uncertainties coherently, laying the groundwork for contemporary surrogate-based robust and reliability design frameworks.

\subsection{Model Uncertainty in Machine Learning Surrogates}

Modern ML methods, particularly those grounded in probabilistic theory, are capable of providing not only point predictions but also estimates of the confidence associated with those predictions. This is essential in engineering design, where downstream decisions rely heavily on the surrogate’s predictive reliability. Among various techniques, \textit{Gaussian Process (GP) regression} is especially popular due to its analytical tractability and ability to return both a predictive mean and standard deviation \cite{rasmussen2006gaussian}. The variance term quantifies model confidence or uncertainty, with larger values indicating higher epistemic uncertainty.

Beyond GP, methods such as Bayesian Neural Networks (BNNs) \cite{neal2012bayesian}, Monte Carlo dropout \cite{gal2016dropout}, and \textit{variational inference for deep learning} \cite{blundell2015weight} have broadened the landscape of uncertainty-aware learning. Ensemble approaches like Random Forests can also offer predictive variability via bootstrap aggregation \cite{breiman2001random}. More recently, \textit{Physics-Informed Neural Networks} (PINNs) \cite{raissi2019physics} and their extensions in UQ \cite{li2023uqml} have gained attention for embedding physical laws directly into the training process, effectively constraining the surrogate to adhere to governing equations and reducing the risk of physically implausible predictions.

In parallel with advancements in ML, the engineering design community has developed rigorous methods to manage uncertainty in simulation-based analysis. While traditional reliability methods have long focused on aleatory uncertainty, the incorporation of epistemic uncertainty into design formulations followed later and has since become an integral part of uncertainty-aware design. Significant progress has been made in quantifying and propagating both types of uncertainty within reliability-based design \cite{sankararaman2013separating,Du_UQ_2024} and robust design frameworks \cite{chen2005,tootchi2025robust}.

Several recent studies \cite{Du_UQ_2024,tootchi2025robust} demonstrate that effective design decisions can be made under joint aleatory and epistemic uncertainties without retraining machine learning models—an important advantage when surrogate models are developed by third parties or embedded in commercial tools. Proposed strategies include both analytical and sampling-based propagation of model uncertainty, incorporation of confidence bounds into design constraints, and the use of hierarchical or multi-fidelity modeling approaches \cite{perdikaris2017nonlinear}.

\subsection{Polynomial Chaos Expansion (PCE)}

Polynomial Chaos Expansion (PCE) is a spectral method widely used in UQ to approximate the response of stochastic systems. It represents a random output variable as a series expansion of orthogonal polynomials defined over standard random variables \cite{xiu2002wiener}. Originating from Wiener’s homogeneous chaos \cite{wiener1938homogeneous}, Generalized Polynomial Chaos (gPCE) extends the original concept to accommodate various input distributions by selecting appropriate polynomial families: Hermite for Gaussian, Legendre for Uniform, Laguerre for Gamma, and Jacobi for Beta distributions.

The stochastic response $Y$, as a function of $D$ independent standard random variables $\mathbf{X} = (X_1, X_2, \dots, X_D)$, is approximated by:
\begin{equation}\label{eq:pce_expansion}
Y(\mathbf{X}) \approx \sum_{k=0}^{P} \alpha_k \Psi_k(\mathbf{X}),
\end{equation}
where:
\begin{itemize}
  \item $\alpha_k$ are deterministic coefficients,
  \item $\Psi_k(\mathbf{X})$ are multivariate orthogonal polynomials formed by products of univariate polynomials, such as $\Psi_{\mathbf{p}}(\mathbf{X}) = \prod_{d=1}^{D} H_{p_d}(X_d)$ for Hermite polynomials,
  \item $P$ is the number of basis terms, determined by the input dimension $D$ and maximum polynomial order $p_{\text{max}}$, with $P+1 = \binom{D + p_{\text{max}}}{p_{\text{max}}}$.
\end{itemize}

A key advantage of PCE is that it enables analytical computation of output statistics when the basis polynomials are orthonormal. For example, with standard normal inputs and Hermite polynomials:
\begin{equation}
\mathbb{E}[Y] = \alpha_0, \qquad \mathrm{Var}[Y] = \sum_{k=1}^{P} \alpha_k^2.
\end{equation}
Several computational strategies have been developed to estimate the PCE coefficients. \textit{Intrusive} approaches such as the stochastic Galerkin method reformulate the governing equations and solve for the coefficients directly, requiring access to and modification of the model \cite{Ghanem1991}. In contrast, \textit{non-intrusive} approaches, including stochastic collocation and regression, treat the model as a black box. Among these, least-squares regression is widely used, where the coefficients are fitted by minimizing the error between simulation outputs and the PCE approximation at selected sample points \cite{Sharma2024, Blatman2011}. Sparse regression techniques such as Least Angle Regression (LAR) \cite{Efron2004} and LASSO \cite{Tibshirani1996} are often applied to reduce the number of basis terms and improve computational efficiency.

PCE has been broadly applied across engineering disciplines, especially in UQ \cite{He2024, yang2023robust, he2023novel,liu2022analytical, zhang2021structural, lee2022reliability, liu2022novel, parnianifard2023expedited, shang2021efficient, he2020adaptive}. In design-oriented tasks, it serves as a surrogate model for simulation-based design, reliability analysis, robust optimization, and global sensitivity analysis \cite{sudret2008global, blatman2011adaptive}. Once constructed, the surrogate enables rapid estimation of output statistics and Sobol' indices \cite{sudret2008global}. PCE is also increasingly used in scientific machine learning and data-driven UQ workflows \cite{SHARMA2024117314,lin2020data}, particularly due to its robustness and ability to incorporate physical constraints. 

Despite these strengths, PCE suffers from the curse of dimensionality: the number of basis terms grows exponentially with input dimension, leading to increased memory and data requirements. Even sparse methods can become inefficient in very high dimensions due to the need to generate the full candidate basis set in advance \cite{luthen2021sparse}.

In summary, PCE is a versatile and effective method for UQ in engineering. Its ability to combine analytical tractability, data efficiency, and physical interpretability makes it well-suited for both engineering analysis and design.

\subsection{Motivation and Contributions for Using PCE in This Work}
This work extends PCE to accommodate model uncertainty. There are several compelling reasons to use PCE. 
\begin{itemize}
    \item \textbf{Unified Framework for Coupled Uncertainty:} PCE provides a natural and unified framework to incorporate both the uncertainty in the input variables and the model itself. By transforming all sources of randomness into a single set of independent standard normal variables, the coupled uncertainty can be propagated through a single PCE model.
    \item \textbf{Computational Efficiency:} Once the PCE coefficients are determined, the mean and standard deviation of the output can be calculated analytically from these coefficients. This avoids the need for a large number of model evaluations typically required by Monte Carlo simulations for accurate moment estimation.
    \item \textbf{Global Sensitivity Analysis:} A significant advantage of PCE, especially with orthonormal basis functions, is its ability to directly compute global sensitivity indices (e.g., Sobol' indices) from the PCE coefficients. This enables the direct quantification of the contribution of each input variable, including the dedicated ML model uncertainty variable (details in the next section), to the total output variance. This provides valuable insights into which sources of uncertainty dominate the overall variability in the prediction.
    \item \textbf{Handling Complex Relationships:} PCE can accurately approximate highly non-linear input-output relationships, provided a sufficiently high polynomial order is chosen and enough training data is available.
    \item \textbf{Foundation for Further Analysis:} Beyond moments and sensitivity indices, a PCE model can be used for other UQ tasks such as the estimation of the complete distribution of the output.
\end{itemize}

The main contributions of this work are threefold. First, we develop a robust PCE-based UQ methodology that captures both aleatory uncertainty in the input variables and epistemic uncertainty in the ML model predictions. These two sources of uncertainty are systematically combined in a unified standard normal space, enabling the construction of a single surrogate model for efficient UQ. Second, we propose a clear framework for global sensitivity analysis based on Sobol' indices that includes ML model uncertainty as an explicit random input. This allows us to quantify the impact of the ML model error relative to physical input variables, offering important guidance on where to focus efforts for reducing uncertainty. Third, we demonstrate the accuracy and efficiency of the proposed method through two examples, comparing the results with large-scale Monte Carlo simulations (MCS). These results show the method's potential for practical engineering applications where reliable and efficient UQ is needed.

It should be noted that this research does not focus on developing ML methods. Instead, it addresses the downstream application of ML in engineering analysis and design, with a particular emphasis on quantifying the effects of uncertainty in ML model predictions on engineering outcomes. The model uncertainty considered in this study originates from the ML models but is not quantified or modeled here; rather, it is treated as a known input to the proposed UQ method. The objective is to evaluate how this model uncertainty, together with traditional input variability, affects predictions for engineering decisions and design.

\section{Methodology}
This section details the proposed method to quantify and propagate coupled uncertainty from both the input variables and the ML model prediction into the final output. The method enables efficient estimation of the output's statistical moments, distribution, and comprehensive sensitivity analysis.

\subsection{Problem Formulation}
As discussed previously, a general computational model $Y = h(\mathbf{X})$ can be replaced by an inexpensive ML model $Y = g(\mathbf{X})$. In many engineering applications, $\mathbf{X}$ is random. The joint Probability Density Function (PDF) of these random inputs is $f_{\mathbf{X}}(\mathbf{x})$. This distribution represents \textit{input uncertainty}, which is \textit{aleatory uncertainty} stemming from inherent variability. The ML model returns a probabilistic response $Y$ for a given input $\mathbf{X}$. This response is characterized by a conditional distribution $F_{Y|\mathbf{X}}(y; \boldsymbol{\theta}(\mathbf{X}))$, where $\boldsymbol{\theta}(\mathbf{X})$ represents the distribution parameters that depend on $\mathbf{X}$. This conditional distribution accounts for \textit{model uncertainty} (also known as \textit{epistemic uncertainty}). This model uncertainty is inherently coupled with the input uncertainty in $\mathbf{X}$. The overall uncertainty containing both types of uncertainties is termed \textit{coupled uncertainty} in this study.

The output $Y$ is thus a random variable due to coupled uncertainty. Let $f_Y(y)$ and $F_Y(y)$ be the PDF and CDF of $Y$, respectively. 

Following the  strategy of decoupling the two types of uncertainty \cite{Du_UQ_2024}, we consider the following transformation:
\begin{equation}\label{eq:prob_integral_transform_Z}
Z = F_{Y|\mathbf{X}}(Y;\boldsymbol{\theta}(\mathbf{X}))
\end{equation}
By the probability integral transform, $Z$ follows a uniform distribution on $[0,1]$, i.e., $Z \sim Unif[0,1]$ and is independent of $\mathbf{X}$ \cite{ditlevsen1996structural}. This auxiliary random variable serves to represent the model uncertainty.  

The predicted response $Y$ by the ML model can then be recovered by inverting this transformation:
\begin{equation}\label{eq:inverse_transform_Y}
Y = F_{Y|\mathbf{X}}^{-1}(Z;\boldsymbol{\theta}(\mathbf{X}))
\end{equation}
This equation indicates that model uncertainty represented by $Z$  and input uncertainty in $\mathbf{X}$ are fully decoupled. The decoupling allows for accounting for model uncertainty with any prediction distribution $F_{Y|\mathbf{X}}(Y;\boldsymbol{\theta}(\mathbf{X}))$.

We now focus on the ML model built using GP regression. For a GP model, the conditional distribution of $Y$ given $\mathbf{X}$ is Gaussian (Normal):
\begin{equation}\label{eq:gp_conditional_dist}
Y|\mathbf{X} \sim \mathcal{N}(M(\mathbf{X}), S^2(\mathbf{X}))
\end{equation}
Here, $M(\mathbf{X})$ is the predictive mean of the GP model, representing the most probable output for a given $\mathbf{X}$, and $S(\mathbf{X})$ is the predictive standard deviation, quantifying the uncertainty or confidence in that prediction. A larger $S(\mathbf{X})$ indicates higher uncertainty, typically in regions where training data is sparse.

From the conditional Gaussian distribution in \eqref{eq:gp_conditional_dist}, the conditional CDF of $Y$ given $\mathbf{X}$ can be expressed as:
\begin{equation}\label{eq:conditional_cdf_phi}
F_{Y|\mathbf{X}}(Y;\mathbf{X}) = \Phi\left[\frac{Y-M(\mathbf{X})}{S(\mathbf{X})}\right] = \Phi(U_Y)
\end{equation}
where $\Phi(\cdot)$ denotes the CDF of the standard normal distribution, and $U_Y = \frac{Y-M(\mathbf{X})}{S(\mathbf{X})}$ is a standard normal random variable.

Comparing \eqref{eq:prob_integral_transform_Z} and \eqref{eq:conditional_cdf_phi}, we obtain:
\begin{equation}\label{eq:y_final_form}
Y = M(\mathbf{X}) + U_Y S(\mathbf{X})
\end{equation}
Thus, the response $Y$ becomes a function of the random input variables $\mathbf{X}$ (which carry aleatory uncertainty) and the auxiliary standard normal variable $U_Y$ (which accounts for model epistemic uncertainty). Both types of uncertainties are now explicitly decoupled, allowing for their joint propagation within a unified framework.

\subsection{Application of PCE}
We now apply PCE and follow the following steps.

\subsubsection{Input Variable Transformation}
We first transform all random input variables into independent standard normal variables.
\begin{itemize}
    \item \textbf{Transformation of $\mathbf{X}$:} Each component $X_i$ of the input vector $\mathbf{X}$ (which may have various distributions) is transformed into a corresponding standard normal variable $U_{X_i}$. This is achieved using the isoprobabilistic transformation:
    \begin{equation}\label{eq:isoprobabilistic_transform_method}
    U_{x_i} = \Phi^{-1}(F_{X_i}(X_i))
    \end{equation}
    where $F_{X_i}$ is the CDF of $X_i$, and $\Phi^{-1}$ is the inverse CDF of the standard normal distribution.  

    \item \textbf{Combined Standard Normal Vector:} The complete set of independent standard normal input variables for the PCE is formed by concatenating the transformed $\mathbf{X}$ variables ($\mathbf{U}_X = (U_{X_1}, \dots, U_{X_n})$) and the model uncertainty variable $U_Y$:
    \begin{equation}\label{eq:combined_U_method}
    \mathbf{U} = (\mathbf{U}_X, U_Y)
    \end{equation}
    If $\mathbf{X}$ has $n$ variables, the total dimension of $\mathbf{U}$ for the PCE is $D = n+1$.
\end{itemize}
This transformation of all random inputs into standard normal variables offers several advantages for PCE:
\begin{itemize}[noitemsep]
    \item \textbf{Orthonormal Basis Selection:} For standard normal variables, Probabilists' Hermite Polynomials form an orthogonal basis with respect to the standard normal probability measure. This orthonormality is crucial for the efficient and stable calculation of PCE coefficients via least-squares regression. It also directly enables the analytical computation of the output's statistical moments (mean and variance) from these coefficients, as shown in \eqref{eq:mean_y_method} and \eqref{eq:variance_y_method}, and, importantly, for global sensitivity analysis. 
    \item \textbf{Unified Representation:} It allows all sources of randomness, regardless of their original distribution (Normal, Lognormal, Uniform, etc.), to be represented in a single, consistent standard normal space. This uniformity simplifies the PCE construction as a single set of multivariate Hermite polynomials can be applied across all dimensions of $\mathbf{U}$.
    \item \textbf{Numerical Stability:} Standardizing the scale and distribution of different input variables can improve the numerical stability and conditioning of the regression problem when solving for the PCE coefficients.
    \item \textbf{Decoupling of Uncertainties:} As established in the problem formulation, this transformation explicitly decouples the aleatory uncertainty (from $\mathbf{X}$) and the epistemic uncertainty (represented by $U_Y$), allowing them to be treated uniformly within the PCE framework.
\end{itemize}

\subsubsection{PCE Expansion}
For an output $Y$ that is a function of the set of independent standard normal random variables $\mathbf{U}$ (as defined in \eqref{eq:combined_U_method}), the PCE approximates $Y$ as a series expansion of orthogonal polynomials:
\begin{equation}\label{eq:pce_expansion_methodology}
Y(\mathbf{U}) \approx \sum_{k=0}^{P} \alpha_k \Psi_k(\mathbf{U})
\end{equation}
where:
\begin{itemize}
    \item $\alpha_k$: Deterministic PCE coefficients that need to be determined.
    \item $\Psi_k(\mathbf{U})$: Multivariate orthogonal Hermite Polynomials that form the basis of the expansion.
\end{itemize}

\subsubsection{Orthogonal Polynomial Basis}
Given that the composite input vector $\mathbf{U}$ consists of independent standard normal variables, Probabilists' Hermite Polynomials are the natural choice for the orthogonal basis functions $\Psi_k(\mathbf{U})$. These polynomials are orthogonal with respect to the standard normal probability density function.
\begin{itemize}
    \item \textbf{Univariate Hermite Polynomials:} The $j$-th order univariate Hermite polynomial $H_j(u)$ is calculated using a recurrence relation:
    \begin{align}\label{eq:hermite_recurrence_method}
    H_0(u) &= 1 \\
    H_1(u) &= u \\
    H_j(u) &= u H_{j-1}(u) - (j-1) H_{j-2}(u) \quad \text{for } j \ge 2.
    \end{align}

    \item \textbf{Multivariate Hermite Polynomials:} A multivariate basis function $\Psi_k(\mathbf{U})$ is constructed as a product of univariate Hermite polynomials, corresponding to a multi-index $\mathbf{p} = (p_1, p_2, \dots, p_D)$:
    \begin{equation}\label{eq:multivariate_hermite_method}
    \Psi_{\mathbf{p}}(\mathbf{U}) = \prod_{d=1}^{D} H_{p_d}(U_d)
    \end{equation}
    The multi-indices are generated such that their sum of powers (total degree) does not exceed a predefined maximum polynomial order. 
\end{itemize}

\subsubsection{Training Data Generation}
To determine the PCE coefficients $\alpha_k$, a set of training data points, consisting of input samples $\mathbf{U}$ and corresponding output samples $Y$, is required. The method for generating these training samples significantly impacts the computational cost and accuracy of the PCE model. We consider three primary strategies for training data generation: Latin Hypercube Sampling (LHS), Tensor-Product Quadrature, and Smolyak Sparse Grid. Note that the training data points are for a ML purpose, they are instead used to determine the PCE coefficients $\alpha_k$.

\textbf{Latin Hypercube Sampling (LHS):}
Latin Hypercube Sampling (LHS) is a non-colliding, stratified Monte Carlo sampling technique commonly used for generating input samples for PCE \cite{mckay1992latin}. It ensures that each stratum of each input variable's probability distribution is sampled exactly once. This stratification leads to a more uniform coverage of the input space compared to simple random sampling, which can significantly improve the efficiency and accuracy of the PCE model, particularly for surrogate modeling of complex, non-linear functions.
\begin{itemize}
    \item \textbf{Advantages:} LHS is relatively simple to implement, provides good space-filling properties, and can be computationally efficient for generating a fixed number of samples $N$. It generally converges faster than simple Monte Carlo method for moment estimation and is robust for a wide range of problems.
    \item \textbf{Disadvantages:} The choice of $N$ is critical and often determined heuristically. While space-filling, it does not inherently exploit the polynomial structure of PCE for optimal placement of points. Its effectiveness can degrade in very high dimensions where even uniform coverage becomes sparse.
    \item \textbf{Application Scope:} Widely used as a general-purpose sampling method for non-intrusive PCE, particularly when the underlying model is expensive and a moderate number of training points are desired. 
\end{itemize}

\textbf{Tensor-Product Quadrature:}
Tensor-product quadrature methods generate a set of collocation points by forming a full Cartesian product of one-dimensional quadrature rules \cite{Xiu2002}. For each input dimension, a specific number of quadrature points are chosen based on the distribution of that random variable (e.g., Gaussian-Hermite points for normal variables, Gauss-Legendre points for uniform variables). The total number of points $N$ is the product of the number of points in each dimension, i.e., $N = \prod_{d=1}^D N_{1D,d}$, where $N_{1D,d}$ is the number of 1D points for dimension $d$. A common choice is to use $p_{\text{max}}+1$ points in each dimension, where $p_{\text{max}}$ is the maximum polynomial order.
\begin{itemize}
    \item \textbf{Advantages:} This method provides highly accurate and theoretically optimal approximations for PCE coefficients for lower-dimensional problems, especially when the function is smooth. The points are specifically chosen to be optimal for numerical integration based on the polynomial basis.
    \item \textbf{Disadvantages:} The primary drawback is the "curse of dimensionality." The number of points grows exponentially with the input dimensionality $D$. This makes it computationally prohibitive for problems with more than a handful of uncertain input variables (typically $D > 5$ to $7$).
    \item \textbf{Application Scope:} Best suited for low-dimensional problems where high accuracy is preferred, and the computational model is not excessively expensive, allowing for the exhaustive evaluation of all collocation points.
\end{itemize}

\textbf{Smolyak Sparse Grid:}
To mitigate the exponential growth of points associated with full tensor-product quadrature in higher dimensions, Smolyak sparse grids offer a more computationally efficient alternative \cite{Smolyak1963, Bungartz2004}. A Smolyak sparse grid constructs a carefully selected subset of the full tensor product grid points, emphasizing points that contribute most significantly to the accuracy of the approximation while reducing redundancy. 
\begin{itemize}
    \item \textbf{Advantages:} Significantly reduces the number of required training points compared to full tensor-product grids, making PCE feasible for higher-dimensional problems where full grids are intractable. It retains theoretical convergence properties while offering substantial computational savings.
    \item \textbf{Disadvantages:} The implementation is more complex than LHS or full tensor products. While much better than full grids, the number of points can still become large for very high dimensions ($D > 20-30$) or very high accuracy requirements.
    \item \textbf{Application Scope:} An excellent choice for medium- to high-dimensional problems (typically $D$ up to 20-30) where the accuracy of quadrature-based methods is desired without the prohibitive cost of full tensor products. It is widely used for non-intrusive PCE in more complex engineering and scientific applications.
\end{itemize}

\textbf{Obtaining $Y$ Samples:}
For each generated input sample $\mathbf{U}^{(j)} = (\mathbf{U}_{x}^{(j)}, U_Y^{(j)})$, regardless of the sampling strategy used, the corresponding output $Y^{(j)}$ is obtained through \eqref{eq:inverse_transform_Y}. Specifically, for each $\mathbf{U}^{(j)}$:
\begin{enumerate}
    \item The $\mathbf{U}_x^{(j)}$ components are transformed back to the original physical space $\mathbf{X}^{(j)}$ using the provided distribution information and \eqref{eq:isoprobabilistic_transform_method}.
    \item The $Y^{(j)}$ sample is calculated using \eqref{eq:inverse_transform_Y}; or \eqref{eq:y_final_form} for a GP ML model, and the equation is given by $Y^{(j)} = M(\mathbf{X}^{(j)}) + U_Y^{(j)}$ .
\end{enumerate}

\subsubsection{PCE Coefficient Estimation}
Once the training data $(\mathbf{U}^{(j)}, Y^{(j)})$ for $j=1, \dots, N$ are obtained, the PCE coefficients $\alpha_k$ are estimated using a least-squares regression approach. This system can be concisely expressed in matrix form:

\begin{equation}\label{eq:matrix_form_method}
\mathbf{Y} = \mathbf{\Psi} \mathbf{\alpha} + \mathbf{\epsilon}
\end{equation}

where:
\begin{itemize}
    \item $\mathbf{Y}$ is an $N \times 1$ vector containing the observed output samples $Y^{(j)}$.
    \item $\mathbf{\Psi}$ is the $N \times (P+1)$ design matrix, where each row $j$ corresponds to the evaluation of all $P+1$ multivariate orthogonal polynomial basis functions at the sample $\mathbf{U}^{(j)}$.
    \item $\mathbf{\alpha}$ is the $(P+1) \times 1$ vector of unknown PCE coefficients.
    \item $\mathbf{\epsilon}$ is the $N \times 1$ vector of residuals.
\end{itemize}
The coefficients $\mathbf{\alpha}$ are then determined by solving this linear system in a least-squares sense:

\begin{equation}\label{eq:least_squares_solution_method}
\mathbf{\alpha} = (\mathbf{\Psi}^T \mathbf{\Psi})^{-1} \mathbf{\Psi}^T \mathbf{Y}
\end{equation}

In numerical implementations, this solution is efficiently computed using a matrix division operator.
\subsubsection{Statistical Moment Calculation}
A significant advantage of using orthonormal polynomial bases for PCE is the ability to directly compute the statistical moments of the output from the estimated coefficients, without further simulations.
\begin{itemize}
    \item \textbf{Mean of Y:} The mean of the output $Y$ is given by the first PCE coefficient, $\alpha_0$, which corresponds to the constant term (where all polynomial orders are zero, $\Psi_0(\mathbf{U}) = 1$):
    \begin{equation}\label{eq:mean_y_method}
    E[Y] \approx \alpha_0
    \end{equation}

    \item \textbf{Standard Deviation of Y:} The variance of the output $Y$ is the sum of the squares of all PCE coefficients, excluding the first one:
    \begin{equation}\label{eq:variance_y_method}
    Var[Y] \approx \sum_{k=1}^{P} \alpha_k^2
    \end{equation}
    This property holds due to the orthonormality of the Hermite polynomials with respect to the standard normal probability measure. The standard deviation is simply the square root of the variance:
    \begin{equation}\label{eq:std_y_method}
    \sigma_Y = \sqrt{Var[Y]}
    \end{equation}

    \item \textbf{CDF or PDF of Y:} There are two approaches. One approach is to use MCS to estimate the CDF or PDF. The other approach is to find higher moments of $Y$, and then use Saddlepoint Approximation to estimate the CDF or PDF. Details are given in \cite{Du_SPA_2006}.
\end{itemize}

\subsubsection{Quantifying the Contribution of Uncertainties: Sensitivity Analysis via PCE}
One of the most powerful features of the PCE framework is the ability to perform global sensitivity analysis directly from the computed PCE coefficients. This allows for the quantification of the contribution of each individual input variable, including the dedicated model uncertainty variable $U_Y$, to the total variance of the output $Y$.

The method employed for this purpose is variance-based sensitivity analysis, typically quantified using \textbf{Sobol' indices}. For a PCE surrogate, Sobol' indices can be calculated analytically from the coefficients $\alpha_k$.

\begin{itemize}
    \item \textbf{First-Order Sobol' Index ($S_i$):} The first-order Sobol' index for an input variable $U_i$ (where $U_i$ can be any $U_{x_j}$ or $U_Y$) quantifies the proportion of the output variance that is directly attributable to the variability in $U_i$ alone, without considering any interactions with other variables.
    It is computed by summing the squares of all PCE coefficients $\alpha_k$ whose corresponding multivariate polynomial $\Psi_k(\mathbf{U})$ depends \textit{only} on $U_i$.
    \begin{equation}\label{eq:sobol_first_order}
    S_i = \frac{1}{Var[Y]} \sum_{k \in \mathcal{I}_i} \alpha_k^2
    \end{equation}
    where $\mathcal{I}_i$ is the set of indices $k$ such that $\Psi_k(\mathbf{U})$ depends only on $U_i$.

    \item \textbf{Total-Order Sobol' Index ($S_{T_i}$):} The total-order Sobol' index for an input variable $U_i$ quantifies the proportion of the output variance attributable to $U_i$, including its direct effect and all its interactions with other input variables. This is a more comprehensive measure of influence.
    It is computed by summing the squares of all PCE coefficients $\alpha_k$ whose corresponding multivariate polynomial $\Psi_k(\mathbf{U})$ depends on $U_i$, regardless of whether it also depends on other variables.
    \begin{equation}\label{eq:sobol_total_order}
    S_{T_i} = \frac{1}{Var[Y]} \sum_{k \in \mathcal{J}_i} \alpha_k^2
    \end{equation}
    where $\mathcal{J}_i$ is the set of indices $k$ such that $\Psi_k(\mathbf{U})$ includes $U_i$ in its functional form. This can also be defined as $S_{T_i} = 1 - S_{\sim i}$, where $S_{\sim i}$ is the first-order index of all variables except $U_i$.

\end{itemize}
By calculating these indices for $U_Y$, we can precisely measure the contribution of the ML model uncertainty to the overall output variance, and consequently, its impact on the output standard deviation. This provides crucial insight into the trustworthiness of the ML surrogate and helps identify whether input uncertainty or model uncertainty is the dominant source of variability. The sum of all total-order Sobol' indices is generally greater than 1 due to interactions, while the sum of first-order indices is less than or equal to 1. The difference between the total and first-order index for a variable indicates the strength of its interaction effects.

\subsection{Implementation Flowchart}
The implementation of the PCE methodology follows a structured workflow, as depicted below:

\begin{enumerate}[label=\textbf{Step \arabic*.}]
    \item \textbf{Define Problem Parameters (Input for the proposed method):} 
    \begin{itemize}
        \item Specify the distributions and parameters for the random input variables $\mathbf{X}$.
        \item Obtain a GP ML model, including mean $M(\mathbf{X})$ and standard deviation $S(\mathbf{X})$. 
        \item Set the number of training points $N$ for PCE.
    \end{itemize}
    \item \textbf{Generate Standard Normal Samples ($\mathbf{U}$):}
    \begin{itemize}
        \item Determine the total number of standard normal variables $D = n + 1$ (for $n$ variables in $\mathbf{X}$ and one $U_Y$).
        \item Use Latin Hypercube Sampling to generate $N$ samples from a uniform distribution in $D$ dimensions.
    \end{itemize}
    \item \textbf{Obtain Output Samples ($Y$):}
    \begin{itemize}
        \item For each generated sample $\mathbf{U}^{(j)}$, transform it into $S(\mathbf{X}^{(j)})$ using its distribution. 
        \item Compute $M(\mathbf{X}^{(j)})$ and $S(\mathbf{X}^{(j)})$, and then calculate $Y^{(j)} = M(\mathbf{X}^{(j)}) + U_Y^{(j)} S(\mathbf{X}^{(j)})$ as in \eqref{eq:y_final_form}.
        \item Store these $Y^{(j)}$ values.
    \end{itemize}
    \item \textbf{Build PCE Surrogate Model:}
    \begin{itemize}
        \item Define the maximum polynomial order (typically 2 for initial analysis).
        \item Construct the design matrix $\mathbf{\Psi}$ by evaluating the multivariate Hermite polynomial basis functions at each sample in $\mathbf{U}$.
        \item Estimate the PCE coefficients $\mathbf{\alpha}$ by performing least-squares regression, as shown in \eqref{eq:least_squares_solution_method}.
    \end{itemize}
    \item \textbf{Calculate Statistical Moments and Sensitivity Indices:}
    \begin{itemize}
        \item The mean of $Y$ is directly given by the first PCE coefficient: $E[Y] = \alpha_0$ \eqref{eq:mean_y_method}.
        \item The variance of $Y$ is the sum of the squares of all other PCE coefficients: $Var[Y] = \sum_{k=1}^{P} \alpha_k^2$ \eqref{eq:variance_y_method}.
        \item The standard deviation of $Y$ is the square root of the variance: $\sigma_Y = \sqrt{Var[Y]}$ \eqref{eq:std_y_method}.
        \item Global sensitivity indices (first-order and total-order Sobol' indices) for each input variable $U_{x_j}$ and for $U_Y$ are computed directly from the PCE coefficients using \eqref{eq:sobol_first_order} and \eqref{eq:sobol_total_order}.
    \end{itemize}
\end{enumerate}

\section{Examples} 

To demonstrate the effectiveness and accuracy of the proposed PCE method, we present two examples. In both cases, the accuracy of the PCE method is compared against results obtained from Monte Carlo Simulations (MCS). While computational efficiency is often a primary driver for using surrogate models and PCE, for these examples, the emphasis remains on the accuracy of the UQ results, particularly since the underlying ML models are computationally inexpensive to evaluate.

\subsection{Example 1: Speed Reducer Shaft}

to show the effectiveness of considering model uncertainty. The shaft is subjected to a random force $F$ and a random torque $T$. The model defines the design margin for yielding failure, which is the difference between the yield strength $S_y$ and the maximum equivalent stress. 
\begin{equation}
    Y = S_y - \frac{16}{\pi d^3}\sqrt{4F^2l^2 + 3T^2}
    \label{eq:yielding_model_expanded}
\end{equation}

This analytical model is inexpensive and allows us to study the proposed method for the case where the responding ML model is accurate with a sufficiently large set of training points. There are five independent random input variables in $\mathbf{X}$. Their distributions and parameters are described in Table \ref{tab:speed_reducer_variables}.

\begin{table}[H]
	\centering
	\caption{Distributions of the variables in speed reducer shaft problem}
	\label{tab:speed_reducer_variables}
	\begin{tabular}{llccc}
		\toprule
		\textbf{Variables} & \textbf{Symbol} & \textbf{Distribution} & \textbf{Mean} & \textbf{Standard Deviation} \\
		\midrule
		Yield strength $S_y$& $X_1$ & Normal & 250 MPa& 30 MPa\\
		Diameter $d$& $X_2$ & Normal & 40 mm& 0.0001 mm\\
		Length $l$& $X_3$ & Normal & 400 mm& 0.0001 mm\\
		Random Force $F$& $X_4$ & Lognormal & 1780 N& 363 N\\
		Random Torque $T$& $X_5$ & Extreme Value Type 1 & 430 N$\cdot$m& 40 N$\cdot$m\\
		\bottomrule
	\end{tabular}
\end{table}

A GP ML model is trained using 100 training points, based on which a UQ is performed. The proposed PCE analysis is conducted using three sampling methods for generating training points: Latin Hypercube Sampling (LHS) with $N=80$ points, Tensor-Product with $N=729$ points, and Smolyak Sparse Grid with $N=13$ points. The maximum polynomial order for all PCEs is set to 2. The number of points is determined by the maximum polynomial order and the number of input variables.

The results obtained from the PCE analyses are compared against a large-scale MCS with $N=100,000$ samples, which serves as a reference solution.

\subsubsection{Results} 
The predicted mean and standard deviation of $Y$ from the different PCE analyses and MCS are summarized in Table \ref{tab:results}.

\begin{table}[H]
	\centering
	\caption{PCE and MCS Results for Mean and Standard Deviation of $Y$}
	\label{tab:results}
	\begin{tabular}{lccc}
		\toprule
		\textbf{Method} & \textbf{Mean($Y$, MPa) (Err\%)} & \textbf{Std($Y$, MPa) (Err\%)} & \textbf{\# Points} \\
		\midrule
		LHS PCE & 124.33 (0.1\%) & 23.34 (0.5\%) & 80 \\
		Tensor-Product PCE & 124.12 (0.0\%) & 23.16 (1.3\%) & 729 \\
		Smolyak PCE & 123.96 (0.2\%) & 23.59 (0.6\%) & 13 \\
		\midrule
		Monte Carlo Sim. & 124.18 & 23.45 & 100000 \\
		\bottomrule
	\end{tabular}
\end{table}

\subsubsection{Sensitivity Analysis} 
\begin{figure}[H]
	\centering
	\includegraphics[width=0.75\textwidth]{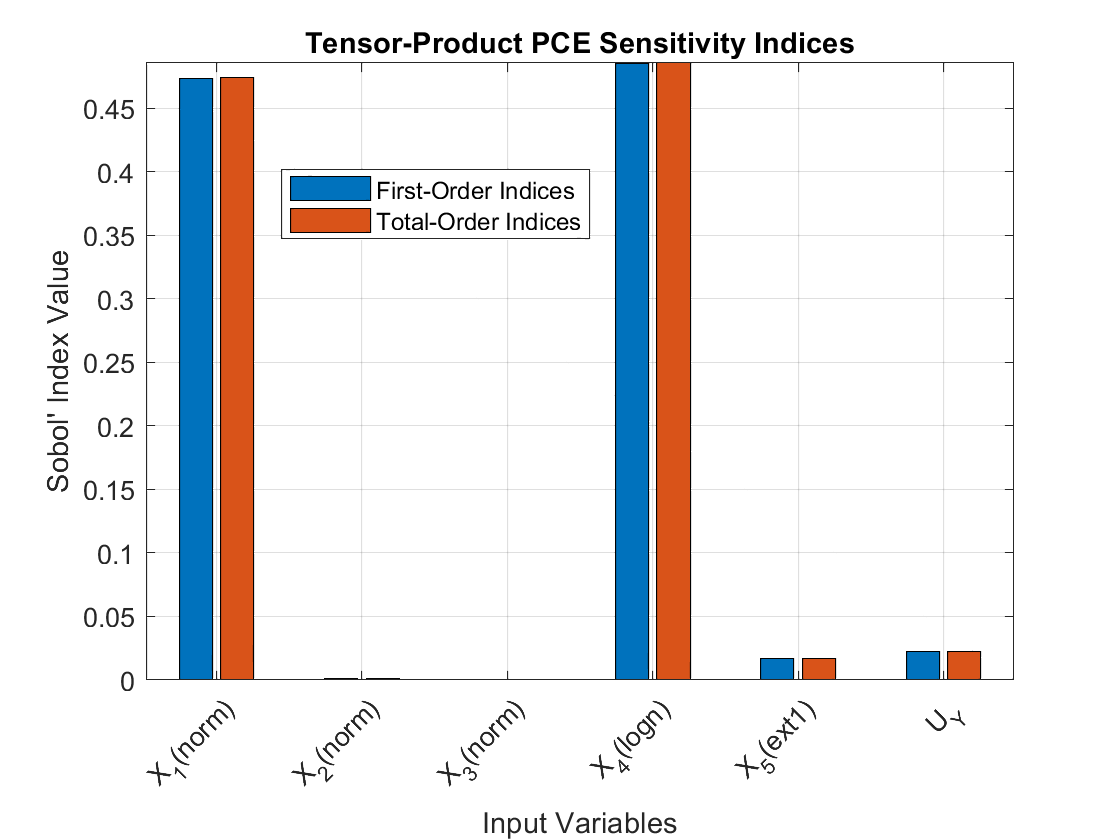}
	\caption{Sensitivity Analysis Results for the ML Model with 100 Training Points (Moderate Model Uncertainty)}
	\label{fig:sobol_100}
\end{figure}

The Sobol' sensitivity indices for each input variable, including the model uncertainty ($U_Y$), are calculated for all three PCE methods. These indices quantify the contribution of each input variable to the total variance of the output $Y$. Since the results are similar from the three methods, the results from only Tensor-Product PCE are provided in Table \ref{tab:sobol_tensor_product} and in Figure \ref{fig:sobol_100}.

\begin{table}[H]
	\centering
	\caption{Sobol' Sensitivity Indices for Tensor-Product PCE}
	\label{tab:sobol_tensor_product}
	\begin{tabular}{lcc}
		\toprule
		\textbf{Variable} & \textbf{First-Order} & \textbf{Total-Order} \\
		\midrule
		$X_1$ (norm) & 0.3918 & 0.3927 \\
		$X_2$ (norm) & 0.0019 & 0.0024 \\
		$X_3$ (norm) & 0.0001 & 0.0006 \\
		$X_4$ (logn) & 0.4736 & 0.4743 \\
		$X_5$ (ext1) & 0.0116 & 0.0121 \\
		$U_Y$        & 0.1192 & 0.1196 \\
		\midrule
		\multicolumn{2}{l}{\textbf{Sum of First-Order:}} & \textbf{0.9983} \\
		\bottomrule
	\end{tabular}
\end{table}

\subsubsection{Discussions} 
As observed from the results in Table \ref{tab:results}, the mean and standard deviation predicted by all three PCE methods are in excellent agreement with those obtained from MCS.

\begin{itemize}
	\item The predicted mean values from LHS PCE (124.33 MPa), Tensor-Product PCE (124.12 MPa), and Smolyak PCE (123.96 MPa) are all close to the MCS result (124.18 MPa), with relative errors ranging from 0.0\% to 0.2\%.
	
	\item Likewise, the predicted standard deviation values from LHS PCE (23.34 MPa), Tensor-Product PCE (23.16 MPa), and Smolyak PCE (23.59 MPa) are consistent with the MCS result (23.45 MPa), showing relative errors between 0.5\% and 1.3\%.
\end{itemize}

This high level of accuracy demonstrates the capability of the PCE method to effectively quantify coupled uncertainty from both the input variables and the ML model's prediction errors, even with a relatively small number of training points compared to MCS ($N=100{,}000$). Notably, the Smolyak PCE achieves comparable accuracy with significantly fewer training points ($N=13$) compared to LHS PCE ($N=80$) and Tensor-Product PCE ($N=729$), highlighting its computational efficiency advantage for higher-dimensional problems.

The sensitivity analysis results (Table \ref{tab:sobol_tensor_product}) provide insight into the contribution of each input variable to the total variance of the output as shown in Figure \ref{fig:sobol_100}. In all PCE methods, $X_1$ (Yield strength) and $X_4$ (Random Force) are consistently identified as the most influential variables, while the contribution from $U_Y$ (ML model uncertainty) is relatively small, indicating a robust ML model. 

\begin{figure}[H]
	\centering
	\includegraphics[width=0.75\textwidth]{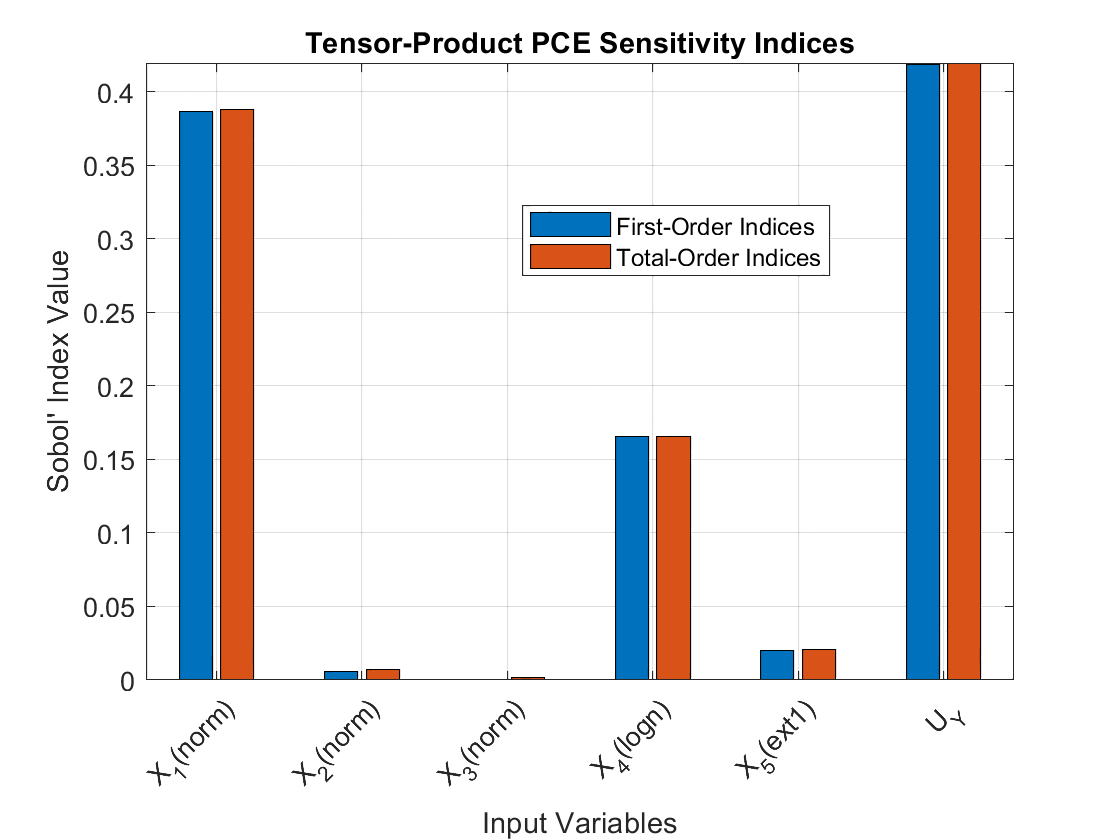}
	\caption{Sensitivity Analysis Results for the ML Model with 100 Training Points (Significant Model Uncertainty)}
	\label{fig:sobol_30}
\end{figure}

\begin{figure}[H]
	\centering
	\includegraphics[width=0.75\textwidth]{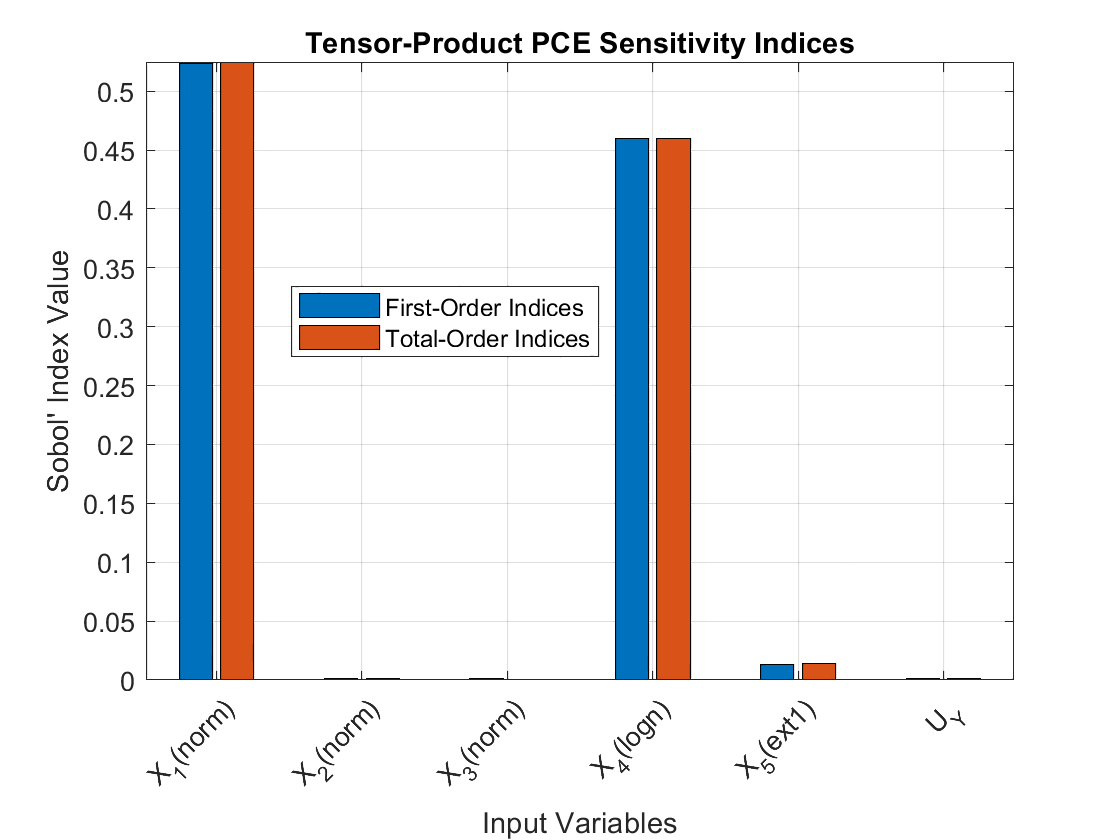}
	\caption{Sensitivity Analysis Results for the ML Model with 500 Training Points (Insignificant Model Uncertainty)}
	\label{fig:sobol_500}
\end{figure}

We now show the impact of model uncertainty by using different sizes of training points. The sensitivity analysis results for ML models with 30 and 500 training points are obtained and are shown in Figure \ref{fig:sobol_30} and Figure \ref{fig:sobol_500}, respectively. The contribution from $U_Y$ (ML model uncertainty) is significant when the number of training points is 30, which is not sufficient as indicated in \ref{fig:sobol_30}. This is due to a large model prediction error. However, the model error becomes low with a large number of training points of 500. Then the contribution of model uncertainty is negligible as indicated in Figure \ref{fig:sobol_500}.

We now examine the effect of model uncertainty by varying the size of the training dataset. Sensitivity analysis results for ML models trained with 30 and 500 points are presented in Figure~\ref{fig:sobol_30} and Figure~\ref{fig:sobol_500}, respectively. When only 30 training points are used, the contribution from $U_Y$ (representing ML model uncertainty) is substantial, as seen in Figure~\ref{fig:sobol_30}, indicating a significant prediction error due to insufficient training. In contrast, with 500 training points, the model error is greatly reduced, and the contribution of $U_Y$ becomes negligible, as shown in Figure~\ref{fig:sobol_500}.

\subsection{Example 2: Uncertainty Quantification of Nonlinear Heat Transfer in a Thin Plate} 

This section discusses the UQ analysis performed on a nonlinear heat transfer problem in a thin plate. The problem, adapted from a standard MATLAB example \cite{mathworksPDEexample}, does not have an analytical solution. It is solved numerically, and the associated model is therefore a black box. For high efficiency, a GP ML model is built to replace this black-box model, and a UQ analysis is therefore needed to assess the uncertainty of the ML model prediction.

\subsubsection{Problem Description}

\begin{figure}[H]
	\centering
	\includegraphics[width=0.75\textwidth]{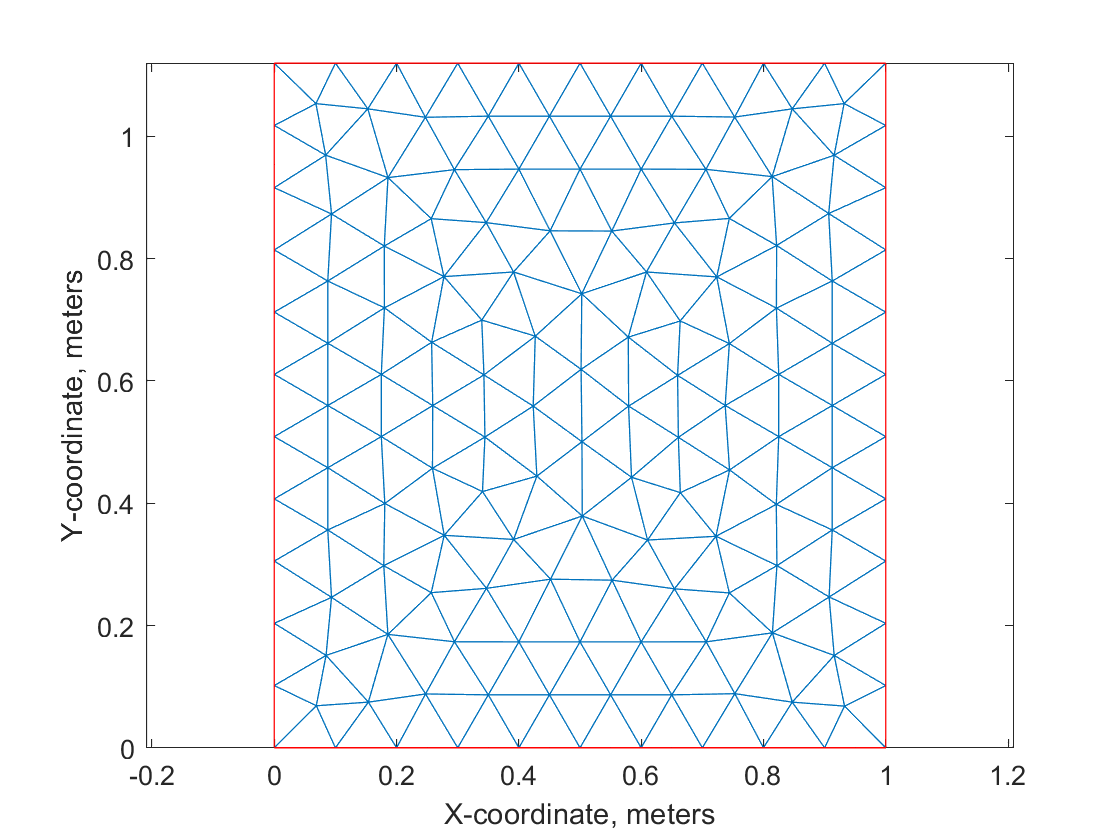}
	\caption{A Thin Plate and Its Mesh}
	\label{fig:mesh}
\end{figure}

The propagation of heat through a thin, rectangular copper plate is modeled considering steady-state conditions. Heat transfer mechanisms include conduction within the plate, convection  and radiation from both plate surfaces to the ambient environment. A fixed temperature boundary condition is applied to one edge of the plate, while other edges are subject to convection and radiation. The governing partial differential equation (PDE) for this system is nonlinear. Given the complexity and nonlinearity of the PDE, a numerical method is employed to determine the temperature distribution across the plate. The output of interest, $t_{top}$, represents the temperature on the top edge of the plate. The plate and its mesh used by the numerical solver are plotted in Figure~\ref{fig:mesh}.

\subsubsection{Random Variables}

Table \ref{tab:random_variables_ex2} summarizes the input random variables, their  units, and their respective probability distributions along with the specified parameters (mean ($\mu$) and standard deviation ($\sigma$)). The ambient temperature, $T_a$, is modeled using an Extreme Type 1 (Gumbel) distribution. The temperature is fundamentally random and time dependent, and the Gumbel distribution is well-suited for modeling the extreme ambient temperature.

\begin{table}[H]
	\centering
	\caption{Distributions of the variables in nonlinear heat transfer problem}
	\label{tab:random_variables_ex2}
	\begin{tabular}{llccc}
		\toprule
		\textbf{Variables} & \textbf{Symbol} & \textbf{Distribution} & \textbf{Mean} & \textbf{Standard Deviation} \\
		\midrule
		Thermal conductivity & $k$ & Normal & 400 W/(m$\cdot$K) & 10 W/(m$\cdot$K) \\
		Convection coefficient & $h_{\text{Coeff}}$ & Normal & 1 W/(m$^2$$\cdot$K) & 0.05 W/(m$^2$$\cdot$K) \\
		Emissivity & $\epsilon$ & Normal & 0.5 & 0.05 \\
		Ambient temperature & $T_a$ & EVT1 & 300 K & 20 K \\
		Height of plate & $H$ & Normal & 1 m & 0.05 m \\
		\bottomrule
	\end{tabular}
\end{table}

\subsubsection{Results}
A GP ML model trained with 200 training points is used for the heat transfer analysis. The UQ analysis is performed using three sampling methods. All approaches use a second-order PCE model. Table \ref{tab:pce_mcs_comparison_ex2} summarizes the predicted mean and standard deviation of $t_{top}$ and compares them with results from MCS using 100{,}000 samples.

\begin{table}[H]
	\centering
	\caption{PCE and MCS Results for Mean and Standard Deviation of $t_{top}$}
	\label{tab:pce_mcs_comparison_ex2}
	\begin{tabular}{lccc}
		\toprule
		\textbf{Method} & \textbf{Mean($t_{top}$, K) (Err\%)} & \textbf{Std($t_{top}$, K) (Err\%)} & \textbf{\# Points} \\
		\midrule
		LHS PCE & $449.41$ (0.0\%) & $20.07$ (0.7\%) & 80 \\
		Tensor-Product PCE & $449.42$ (0.0\%) & $20.39$ (0.9\%) & 729 \\
		Smolyak PCE & $449.23$ (0.0\%) & $19.98$ (1.1\%) & 13 \\
		\midrule
		Monte Carlo Sim. & $449.30$ & $20.21$ & 100000 \\
		\bottomrule
	\end{tabular}
\end{table}

\subsubsection{Sensitivity Analysis}
\begin{figure}[H]
	\centering
	\includegraphics[width=0.75\textwidth]{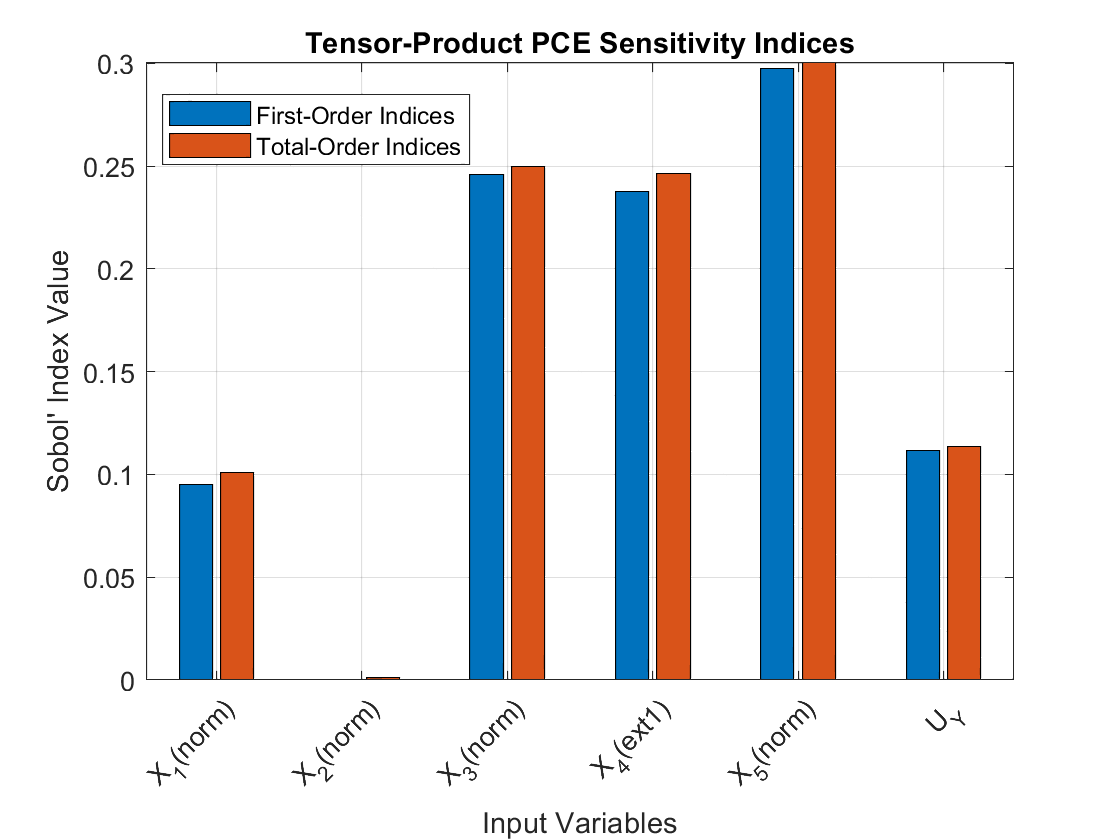}
	\caption{Sensitivity Analysis Results for the ML Model with 200 Training Points}
	\label{fig:sobol_exp2_300}
\end{figure}

To investigate the relative importance of the input variables and model uncertainty, Sobol' sensitivity indices are computed using each PCE model. Both first-order and total-order indices are extracted. Since the results across the three sampling methods are comparable, only those from the Tensor-Product PCE model are included in Table \ref{tab:sobol_tensor_ex2} and Figure \ref{fig:sobol_exp2_300} for illustration.

\begin{table}[H]
	\centering
	\caption{Sobol' Sensitivity Indices for Tensor-Product PCE (Example 2)}
	\label{tab:sobol_tensor_ex2}
	\begin{tabular}{lcc}
		\toprule
		\textbf{Variable} & \textbf{First-Order} & \textbf{Total-Order} \\
		\midrule
		$X_1$ (norm) & 0.0951 & 0.1007 \\
		$X_2$ (norm) & 0.0001 & 0.0011 \\
		$X_3$ (norm) & 0.2460 & 0.2498 \\
		$X_4$ (ext1) & 0.2376 & 0.2465 \\
		$X_5$ (norm) & 0.2974 & 0.3002 \\
		$U_Y$ & 0.1118 & 0.1137 \\
		\midrule
		\multicolumn{2}{l}{\textbf{Sum of First-Order:}} & \textbf{0.9881} \\
		\bottomrule
	\end{tabular}
\end{table}

\subsubsection{Discussions}
The results show that all three PCE methods closely replicate the MCS results for both the mean and standard deviation of the quantity of interest $t_{top}$, with relative errors less than 2.0\%. Notably:

\begin{itemize}
	\item The predicted mean temperatures from all three PCE methods match the MCS value ($449.30$ K) very closely, with errors below 0.1\%.
	\item The standard deviations predicted by LHS PCE ($20.07$ K), Tensor-Product PCE ($20.39$ K), and Smolyak PCE ($19.98$ K) differ from the MCS result ($20.21$ K) by only 0.7\%, 0.9\%, and 1.1\%, respectively.
\end{itemize}

The Sobol' sensitivity analysis (Table \ref{tab:sobol_tensor_ex2} and Figure \ref{fig:sobol_exp2_300}) shows that $X_5$ and $X_3$ are the most influential variables, followed by $X_4$. The contribution from $U_Y$, representing ML model uncertainty, is around 11.2\%, indicating that although the model is reasonably accurate, prediction error still has a measurable impact on the results.

\section{Conclusions}
This paper presents a robust Polynomial Chaos Expansion (PCE) methodology for uncertainty quantification in machine learning (ML) surrogate models, specifically addressing the challenge of simultaneously accounting for uncertainty in input variables and prediction errors from ML regression. The proposed framework effectively integrates these two sources of uncertainty by transforming all random inputs into a unified standard normal space, allowing for the construction of a single PCE surrogate.

A key contribution of this work is not only the efficient propagation of coupled uncertainty for accurate mean and standard deviation estimates, but also the clear framework for global sensitivity analysis using Sobol' indices. By treating the ML model uncertainty as an extra random input, the method allows us to clearly break down the total output variation to each input variable—and importantly—to the ML model’s own uncertainty. This gives valuable insights for engineering analysis. It helps designers see which sources of variation matter most and guides efforts to reduce uncertainty, either by improving the ML model with better data or by tightening control over physical input variables. The examples show that the PCE method is both accurate and efficient compared to Monte Carlo simulation, proving it useful for real engineering problems where reliable predictions are needed.

This research lays a foundation for applying PCE to problems involving coupled uncertainty in machine learning models and its inputs. While three approaches for generating collocation points—Latin Hypercube Sampling, tensor-product quadrature, and Smolyak sparse grid—are demonstrated in the examples, other advanced sampling techniques, particularly for high-dimensional problems, can be explored in future work. Although this study focuses on estimating the mean and standard deviation of the prediction, the results from the proposed method can be readily used to approximate the full probability distribution, for instance, via the Saddlepoint Approximation. Future research may further investigate full distribution estimation using alternative UQ methods in conjunction with the proposed framework. Another promising direction is leveraging the UQ results to improve ML models and to guide engineering design optimization with built-in redundancy, thereby ensuring high reliability and robustness under the coupled uncertainty.    

\section{Conflict of Interest Statement}
The author states that there is no conflict of interest.

\bibliographystyle{unsrt}
\bibliography{references} 

\end{document}